\documentclass{article}

\usepackage{aaai}
\usepackage{xcolor}
\usepackage{hyperref}
\usepackage{comment}

\newcommand{\st}[1]{\textcolor{violet}{Solene: #1}}

\begin{document}

\title{Reducing incompleteness in the game of Bridge using PLP}

\author{Junkang LI,\textsuperscript{1}
Sol\`ene TH\'EPAUT,\textsuperscript{1}
V\'eronique VENTOS,\textsuperscript{1}\\
\textsuperscript{1}{NukkAI Lab, Paris, FRANCE}\\
\href{mailto:junkang.li@nukk.ai}{junkang.li@nukk.ai},
\href{mailto:sthepaut@nukk.ai}{sthepaut@nukk.ai}, 
\href{mailto:vventos@nukk.ai}{vventos@nukk.ai}}

\maketitle
\begin{abstract}
Bridge is a trick-taking card game requiring the ability to evaluate probabilities since it is a game of incomplete information where each player only sees its cards. In order to choose a strategy, a player needs to gather information about the hidden cards in the other players' hand. We present a methodology allowing us to model a part of card playing in Bridge using Probabilistic Logic Programming.
\end{abstract}

\section{Introduction}

Playing Bridge is a very difficult task for both human and computer. From a game-theoretic point of view, Bridge, as a game of incomplete information, is much more complicated than games of complete information such as Chess and Go. More precisely, Bridge is probabilistic, partially observable, sequential and multi-agent. These features present difficulties and interests for AI research. Due to partial observability, the search space for planning is exponentially larger than fully observable cases. In addition, the multi-agent aspect (two players against two, both collaborative and adversarial) makes branching factor larger than in two player games. Therefore, Bridge is an interesting and challenging topic for AI research.

It has already been shown in~\cite{legras2018game} that Bridge is a killer application for Inductive Logic Programming. The goal of this paper is to convey our belief that Bridge is also an ideal test field for both PLP and PILP~\cite{raedt2016statistical,de2008probabilistic,riguzzi2018foundations}.

In the first part of this paper, we briefly introduce Bridge and the probabilistic reasoning in this game. The second part is dedicated to our experiments related to the inference lead problem.

\section{Background}
\paragraph{The game of Bridge in short}

The interested readers can refer for instance to~\cite{mahmood2014bridge} for a more complete presentation of the game of Bridge.

Bridge is a trick-taking card game opposing four players divided in two partnerships (pairs). A standard $52$ card pack is shuffled and each player receives a hand of $13$ cards that is only visible to it. Pairs stand across each other. A Bridge deal is divided into two major playing phases: the bidding phase and the card play.

The goal of the bidding phase is to reach a contract which determines the minimum number of tricks the pair commits to win (between $7$ and $13$) during the card play, either with no trump or with a determined suit as trump. 

During the card play, the goal is to fulfill (for the declarer) or to defeat (for the defenders) the contract reached during the bidding phase.

Bridge robots are still far from human expert level. During the last three years, the World Computer-Bridge Championship has been won by Wbridge5 developed by Yves Costel and boosted by a method presented in~\cite{ictai2017}. One can refer to~\cite{paul2010bethe} for an overview of computer Bridge.

\paragraph{Probabilistic and logic reasoning in Bridge}
In a bridge deal, assuming that players know exactly how the cards are distributed among the four hands will yield a simplified version of bridge which can be easily solved both by bridge experts and by programs such as the Double Dummy Solver developed by Bo Haglund. Hence the difficulty of Bridge is largely due to the fact that the game is partially observable.

In this game  of incomplete information, the main goal of each player consists then in rebuilding the hidden hands with respect to it in order to choose the optimal strategy. Throughout the game, the incompleteness decreases either deterministically (e.g when cards are put on the table) or probabilistically (e.g. by rules related to the bidding or card play conventions). 

Without any additional information, one may assume that the hidden cards are uniformly distributed among the hidden hands. In order to choose a strategy, a player needs to gather as much information as possible, which may be provided by different sources such as the auctions made during the bidding phase (out of the scope of the paper), and the cards played during the card play. Such information influences the probabilities of distributions of the hidden cards with respect to a player and therefore the optimal strategy of the player.

If cards are played at random, then it is not possible to obtain additional information except the position of the played cards. Fortunately for us, and for the interest of the game of Bridge, normal players do not play randomly, but attempt to play in their best interest. To achieve this goal, they follow specific rules and use logical reasoning to choose the optimal strategy with respect to the information they gathered. This behaviour can be exploited by other players to update the probabilities of presence of the remaining cards in the hidden hands, as long as they have insight about the strategy of their partner or opponents.

In conclusion, Bridge is a natural test field for PLP due to its omnipresent probabilistic and logic reasoning mentioned above.

\section{Inference Lead Problem}
\paragraph{Problem setting}
During the bidding step, the incompleteness is maximum with $39$ hidden cards for each player. During the card play, the player on the left of the declarer (called \emph{leader}) leads the first trick of the game. The declarer's partner  (called \emph{Dummy}) then lays his cards face up on the table. The incompleteness is now $25$ hidden cards for the declarer and the leader's partner, and $26$ for the leader.

The lead is usually chosen by the leader according to a set of rules, which may depend on its hand and its knowledge about other players' hand deduced from the bidding. This set of rules, called leading rules, is in theory known by the other players.

Hence the lead conveys valuable information about the leader's hand, available both to his partner and to the declarer. The \emph{inference lead problem} consists in analyzing the lead in order to reduce the incompleteness of a game. Such a probabilistic and logic inference task is particularly important at the beginning of the game, since the uncertainty of the situation is the highest.

\paragraph{Methodology of experiments}
In this paper, we focus on the inference lead problem from the declarer's point of view. We have carried out several experiments\footnote{For more details of the experiments, one can consult the Master thesis \cite{li2019probabilistic}.} on using PLP to model how declarer updates its knowledge of the hidden cards (i.e. its belief states) by exploiting the information conveyed by the lead.

First, we have chosen a simple set of $5$ leading rules that the leader is supposed to follow, written as a PLP program. Since the inference module only uses this program as a black box that takes a leader's hand as input and outputs the lead of this hand according to the rules, any set of rules forming a complete decision tree can be used.

Theoretically speaking, we can assume that after the bidding phase, the declarer sees dummy's hand first and then the lead. The moment just before the lead, there are $26$ hidden cards from the declarer's prospective. Without additional information from the bidding phase, each of these hidden cards has a prior probability of $0.5$ to be in the leader's hand. In other words, they can be described as unbiased Bernoulli random variables. Notice that they are not independent, since each player should have exactly $13$ cards.

Now we need a PLP model to describe the declarer's belief states, i.e. how hidden cards are distributed between the two hidden hands. A simple but elegant approach is to create an unbiased Bernoulli random variable for each card, then impose the evidence that exactly $13$ cards are in the leader's hand. However, in this case the search spaces of PLP is $2^{26}$, which is not prohibitively large but unfeasible since PLP implementations such as ProbLog are extremely slow.

Due to the constraint that each player has $13$ cards, the real search space is significantly smaller than $2^{26}$, a fact that we should exploit. If we are only interested in the hidden cards of the led suit, the search space is even smaller. A better approach consists in factoring the representation of the problem.

We have decided to use one single variable which describes the leader's led suit. The values that this variable, called \emph{hand variable}, can take any possible holdings of the leader in the led suit. Trading $26$ binary variables with $1$ variable that can take $2^n$ different values, where $n$ is the number of hidden cards in the led suit with respect to the declarer, greatly reduces the search space of our PLP model and boosts its performance.

Once the belief states are modelled, the lead can be imposed as an evidence. Then it is straightforward to perform inference task in PLP to model how declarer updates his belief states in terms of the lead he observes and the leading rules according to which the lead is chosen.

\paragraph{Results and conclusion}
Our new PLP model is quite efficient. In fact, with $10$ hidden cards in the led suit, it takes less than $30$ seconds on an ordinary personal computer to update the belief states and to compute the posterior probability of finding each card in the leader's hand. This is fast enough for practical purpose, since in reality we are rarely interested in performing such inference tasks in a suit with more than $10$ hidden cards. With some additional refinement tricks, the program can be optimized even more.

Our approach is general enough to accommodate many further upgrades and new features without changing the PLP program structure or its performance. For example, taking into account information conveyed by the auctions which change the distribution (prior to seeing the lead) of the hand variable. We are able to exploit this information by computing this prior distribution of the hand variable using Python then passing it to the PLP program for lead inference, without degradation of the performance.


\bibliographystyle{aaai}
\bibliography{main}

\end{document}